\documentclass[letterpaper, 10pt, conference]{ieeeconf}
\IEEEoverridecommandlockouts
\usepackage{tikz}
\usepackage{booktabs}
\usepackage{graphicx}
\usepackage{caption}
\usepackage{subcaption}
\usepackage{tikz}
\usepackage{colortbl}
\usepackage{afterpage}
\usepackage{cite}
\usepackage{cuted}
\usepackage{capt-of}
\usepackage{amssymb}
\usepackage{pifont}
\usepackage{xcolor}
\usepackage{trimclip}
\usepackage{hyperref}
\usepackage{balance}
\newlength{\objectsheight}
\newlength{\objectsgap}

\definecolor{lavender}{rgb}{0.9, 0.9, 0.98}
\definecolor{red}{RGB}{255, 0, 0}
\definecolor{orange}{RGB}{252, 130, 62}
\definecolor{blue}{RGB}{0, 0,255}
\definecolor{darkgreen}{RGB}{0, 180,0}
\definecolor{lightgray}{gray}{0.9}
\definecolor{lightblue}{rgb}{0.68,0.85,0.9}
\definecolor{lightorange}{rgb}{1.0, 0.8, 0.6}
\definecolor{myblue}{RGB}{30, 100, 200}

\definecolor{mygreen}{HTML}{34a853}
\definecolor{myred}{HTML}{ea4335}
\newcommand{\ibmtick}{\textcolor{mygreen}{\ding{51}}}
\newcommand{\ibmcross}{\textcolor{myred}{\ding{55}}}

\newcommand{\name}{SoTa}

\newcommand{\rv}{$\mathrm{R}^{\mathrm{V}}$}
\newcommand{\rvt}{$\mathrm{R}^{\mathrm{V},\mathrm{T}}$}
\newcommand{\rvthv}{$\mathrm{R}^{\mathrm{V},\mathrm{T}}+\mathrm{H}^\mathrm{V}$}
\newcommand{\rvthvt}{$\mathrm{R}^{\mathrm{V},\mathrm{T}}+\mathrm{H}^{\mathrm{V},\mathrm{T}}$}
\newcommand{\rvtlvt}{$\mathrm{R}^{\mathrm{V},\mathrm{T}}+\mathrm{L}^{\mathrm{V},\mathrm{T}}$}

\title{\LARGE \bf
\name: Soft Tactile Skins for Dexterous Manipulation
}

\author{Jingyun Yang$^{1*}$, Baiyu Shi$^{1*}$, Timothy Yu$^{1*}$, Haitian Liu$^{2}$, Alberta Longhini$^{1}$, Weichen Wang$^{1}$, \\
Rika Antonova$^{3}$, Zhenan Bao$^{1}$, Jeannette Bohg$^{1}$%
\thanks{$^{*}$ Equal contribution. Correspondence to:}%
\thanks{jingyuny@stanford.edu (learning) and
baiyushi@stanford.edu (sensor).}%
\thanks{$^{1}$Stanford University;
$^{2}$Tsinghua University;
$^{3}$University of Cambridge.}%
}

\begin{document}

\begingroup
\makeatletter
\long\def\@makefntext#1{\noindent#1}
\makeatother
\maketitle
\endgroup

\thispagestyle{empty}
\pagestyle{empty}
\begin{strip}
\vspace{-1.8cm}
    \centering
    \includegraphics[width=\textwidth]{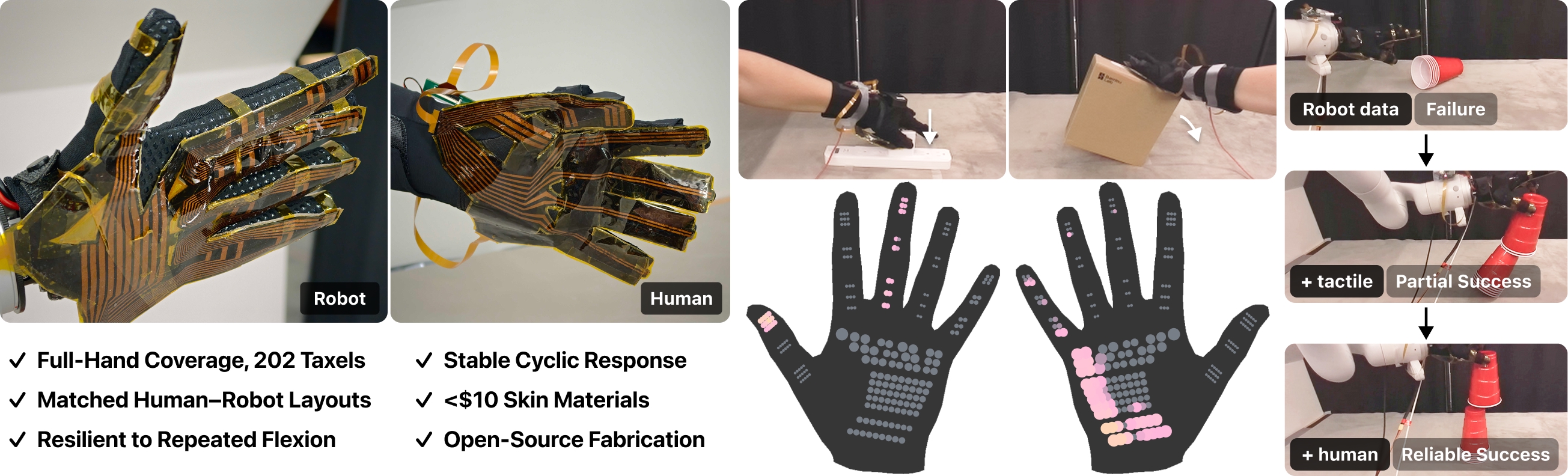}
    \captionsetup{font=footnotesize}
    \captionof{figure}{
    \textbf{The \name~sensor enables dexterous policy learning from human and robot tactile demonstrations.}
    \textbf{Left:} Our soft capacitive skins provide full-hand tactile sensing coverage for human and robot hands.
    \textbf{Middle:} Paired image and tactile sequences illustrate tactile sensor signals in plug insertion and box reorientation tasks.
    \textbf{Right:} Representative dexterous policy rollouts in the cup picking task illustrate the benefits of tactile feedback (top $\to$ middle) and human-robot co-training (middle $\to$ bottom): the robot retrieves a single cup rather than toppling the cups or lifting a nested stack.
    }
    \label{fig:teaser}
    \vspace{-8pt}
\end{strip}

\begin{abstract}
A growing body of work suggests that tactile sensing gives robot policies contact information that complements vision in dexterous manipulation. However, visuo-tactile robot data remains scarce: dexterous demonstrations require teleoperating robots, which limits dataset scale. Human demonstrations are far cheaper to collect and offer a path to scale this data, but only if human and robot hands carry tactile sensors with corresponding signals. This requires sensors that conform to different hand geometries, cover the full hand, and share a common layout across embodiments.
We present \name, a low-cost capacitive tactile skin that provides full-hand coverage on humans and robots while preserving a shared layout of 202 taxels across corresponding finger and palm regions. Our multilayer design with fabric electrodes enables in-house fabrication of thin, soft skins with customizable geometry for under \$10 in materials per skin.
The sensor retains over 97\% of its initial response span after 10,000 loading-unloading cycles with traces retaining continuity through 1,280 tight-fist folding cycles.
The shared taxel layout supports human-robot co-training with a common tactile encoder and no learned cross-sensor mapping.
Across three contact-rich manipulation tasks, tactile observations improve in-distribution success over vision-only policies.
With a fixed robot demonstration budget, adding human demonstrations more than doubles mean success across eight evaluation conditions, from 22.8\% to 45.9\%, improving success in all five out-of-distribution conditions.
We plan to open-source the resources needed to fabricate and operate these skins.
Project website: \href{https://sota-skin.github.io/}{\texttt{https://sota-skin.github.io}}
\end{abstract}

\section{Introduction}

Many dexterous manipulation tasks, such as lifting a fragile cup, wiping dirt from a table, or inserting a charger into a power strip, require carefully controlled contact between objects and different parts of the hand.
Vision-based policies can perform a wide range of manipulation tasks~\cite{bridgedatav2,droid2024,openxembodiment2023,
pi02024,aloha2023,mobilealoha2024}, but they observe contact only indirectly, and a growing body of work suggests that tactile or force observations improve performance on contact-rich tasks~\cite{rotateit2023,yuan2024robosyn,pattabiraman2024,vitac2024,lin2025learning,chen2025dexforceextractingforceinformedactions,yu2025forcevla,tavla2025,adeniji2025feel}.

Scaling visuo-tactile policy learning, however, is limited by data. Dexterous robot demonstrations require teleoperation~\cite{dexpilot2020, qin2023anyteleop, bidexteleop2024}, which is slow and depends on costly hardware. Human demonstrations are much cheaper to collect and could supplement robot data, but only if human hands carry tactile sensors whose signals correspond to the robot's. Existing visuo-tactile systems rarely meet this requirement. Most sense contact only at the fingertips~\cite{rotateit2023,suresh2024neuralfeels,pattabiraman2024, lin2025learning}, so they miss contacts on the phalanges and palm in tasks such as table wiping and plug insertion, and most are built for a single embodiment. A sensor that supports learning from both humans and robots must therefore cover the full hand, conform to different hand geometries, withstand repeated contact and flexing, and present the same tactile structure on both embodiments.

We present \name, a low-cost capacitive tactile skin that meets these requirements.
Our laser-patterned, multilayer fabric-electrode design
produces thin, soft skins whose geometry is fitted to each hand while preserving a shared layout of 202 taxels, with corresponding indices assigned to the same semantic hand regions. Each skin costs under \$10 in materials, responds to forces from 20mN to 10N, retains over 97\% of its initial response span after 10,000 loading–unloading cycles, and keeps trace continuity through 1,280 tight-fist folding cycles.
Because both embodiments share the same layout, human and robot observations use a common tactile encoder without a learned cross-sensor mapping. Human demonstrations contain no action labels (as robot demonstrations do), so we incorporate human demonstrations through an auxiliary objective that predicts task phases from human and robot visuo-tactile observations, while action supervision comes only from robot data.

We evaluate the system on three dexterous manipulation tasks involving contact with different parts of the hand: box reorientation, cup picking, and plug insertion. Tactile observations improve in-distribution success on all three tasks, although these benefits do not consistently extend to out-of-distribution conditions. 
With a fixed robot-demonstration budget, adding human visuo-tactile demonstrations improves every evaluated condition relative to robot-only visuo-tactile training, raising average success from 22.8\% to 45.9\%.

Our contributions are:
(1) a soft, flexible, low-cost, and customizable capacitive tactile skin with coverage across the fingers and palm and a shared 202-taxel layout for human and robot hands; (2) characterization of the sensor's response, cyclic stability, flexing durability, and motion-induced artifacts; and (3) a language-supervised training pipeline that incorporates human visuo-tactile demonstrations into robot policy learning, with real robot experiments demonstrating improvements over robot-only training.
We will release the resources needed to fabricate and operate these skins.
\begin{table*}[t]
\vspace{4pt}
\captionsetup{font=footnotesize}
\centering
\caption{\textbf{Comparison of tactile interfaces relevant to SoTa.}
The upper block lists \textbf{robot-mounted tactile skins} (without human-side wearables).
The lower block lists \textbf{wearable interfaces with a human side} (human-only or bridging human and robot embodiments).
In each block, prior systems are chronologically ordered.
\textbf{Coverage}: FT~=~fingertip, FT+P~=~fingertips and palm, FH~=~full hand (all phalanges + palm).
\textbf{Embodiment}: H~=~worn by human, R~=~mounted on robot, H+R~=~both.
\textbf{Aligned H/R}: HW~=~hardware/layout aligned, SW~=~learned alignment, --~=~not addressed.
\textbf{Custom fit}: per-hand physical customization vs.\ one-size-fits-all.}
\label{tab:glove-comparison}
\resizebox{\textwidth}{!}{%
\begin{tabular}{lllllll}
\toprule
\textbf{System} & \textbf{Sensor} & \textbf{Coverage} & \textbf{Sensing} & \textbf{Embod.} & \textbf{Aligned} & \textbf{Custom} \\
                &                  &                    & \textbf{elements} (total)         &                  & \textbf{H/R}     & \textbf{fit}    \\
\midrule
\multicolumn{7}{l}{\textit{Robot-mounted tactile skins (no human-side wearable)}} \\
\midrule
ReSkin, AnySkin \cite{reskin2021, anyskin2024}                  & Magnetic              & Patch       & 5/patch & R   & --  & no (one-size)           \\
AllTheFeels \cite{allthefeels2023}        & Magnetic              & FT+P  & 56     & R   & --  & no (one-size)           \\
DexSkin \cite{dexskin2025}                & Capacitive            & Round Finger & 120    & R   & --  & yes (per-gripper)   \\
Sharma \cite{sharma2025selfsupervised}  & uSkin (Magnetic)      & FH & 368          & R   & --  & no (one-size)          \\
\midrule
\multicolumn{7}{l}{\textit{Wearable interfaces with a human side (human-only or human + robot)}} \\
\midrule
STAG \cite{sundaram2019learning}          & Piezoresistive (Knit) & FH          & 548(H)          & H   & --  & no (one-size)            \\
MimicTouch \cite{yu2024mimictouch}        & GelSight Mini (Vision)& FT (single) & 1 Sensor            & H+R & HW   & no (one-size)          \\
FlexGlove \cite{murphy2025fits}           & Piezoresistive (FPCB) & FH          & 167(H)    & H   & --  & yes (per-user) \\
DexUMI \cite{dexumi2025}                  & Sensor-matched (FSR / EM) & FT      & sensor-matched & H+R& HW (rigid exo)& yes (per-robot) \\
DexOp \cite{fang2025dexop}                & GelSight (Vision)& FH (FT+P+phal.)& multi-cam & H+R & HW (linkage)  & yes (per-robot) \\
Feel-the-Force \cite{adeniji2025feel}     & AnySkin (Magnetic)    & FT (thumb)  & 5(H) + 5(R) & H+R & SW (force pred.) & no (one-size) \\
OSMO \cite{osmo2025}                      & Magnetic              & FT+P        & 12(H) + 12(R)  & H+R & HW (same glove)& no (one-size) \\
OpenTouch \cite{opentouch2025}            & Piezoresistive (FPCB)  & FH          & 169(H)          & H   & --  & no (one-size) \\
UniTacHand \cite{unitachand} & Heterogeneous tactile & FH & 137(H) + 1062(R) & H+R & SW (MANO + latent) & no (one-size) \\
TactAlign \cite{tactalign2026}            & Magnetic (heterog.)   & FT          & 5(H) + 120(R) & H+R & SW (rect.\ flow)& no (one-size)            \\
\midrule
\textbf{SoTa (ours)} & \textbf{Capacitive (fabric elec.)} & \textbf{FH} & \textbf{202(H) + 202(R)} & \textbf{H+R} & \textbf{HW (shared layout)} & \textbf{yes (per-hand)} \\
\bottomrule
\end{tabular}%
}
\vspace{-12pt}
\end{table*}

\section{Related Work}

\textbf{Visuo-tactile and force-aware policy learning.}
A growing body of work suggests that combining vision with touch or force can improve precision and sample efficiency on contact-rich tasks, both at execution time and as a representation-learning signal \cite{sferrazza2024power,vitac2024,rotateit2023,yuan2024robosyn,suresh2024neuralfeels,chen2025dexforceextractingforceinformedactions,parésmorlans2026demystifyingvlasfailcontactrich,lin2025learning}. Most visuo-tactile policies, however, rely on localized or sparse sensing. Work on multi-fingered hands \cite{lin2025learning,rotateit2023,yuan2024robosyn} has used fingertip sensors \cite{rotateit2023,lin2025learning} or a small set of force-sensing resistors on the palm and finger links~\cite{yuan2024robosyn}, while gripper-based systems use sparse magnetic patches on parallel-jaw grippers \cite{pattabiraman2024}, or dense arrays on soft parallel-jaw fingers \cite{vitac2024}. A parallel line of work injects force or torque signals directly into vision-language-action models \cite{yu2025forcevla,tavla2025,parésmorlans2026demystifyingvlasfailcontactrich} or force-conditioned imitation learning \cite{chen2025dexforceextractingforceinformedactions,adeniji2025feel}.
Most of these approaches are trained on robot data, which is costly to collect for dexterous hands. Methods that draw on human demonstrations do so with limited or heterogeneous sensing: \cite{adeniji2025feel} measures force at a single human fingertip, and FTP-1~\cite{ftp} pretrains a generalist tactile policy on human and robot data from heterogeneous sensors, using sensor-specific encoders.
We build on FTP-1 with a tactile sensing interface that provides full-hand coverage and a shared taxel layout across human and robot hands, so that human and robot demonstrations can share a single tactile encoder without a separate cross-sensor alignment stage.

\textbf{Tactile interfaces for dexterous manipulation.}
Tactile sensing for dexterous manipulation has advanced through both robot-mountable tactile skins and human-worn tactile gloves.
Robot skins use magnetic
\cite{reskin2021,anyskin2024,pattabiraman2024,allthefeels2023}
and capacitive~\cite{dexskin2025} sensing, 
with coverage ranging from localized patches and fingertips to the full hand~\cite{sharma2025selfsupervised}. Human-worn systems such as STAG~\cite{sundaram2019learning}, FlexGlove~\cite{murphy2025fits} or OpenTouch~\cite{opentouch2025} provide full-hand tactile sensing, and FlexGlove additionally customizes sensor geometry to each user.
These systems, however, are designed for a single embodiment. Interfaces designed for both embodiments use mechanically matched structures~\cite{dexumi2025,fang2025dexop} or identical sensors~\cite{osmo2025,yu2024mimictouch}, but they constrain the wearer to robot-specific hardware or provide coverage limited to the fingertips and palm.
\name~instead fits a thin, flexible, and low-cost skin to each hand while preserving a shared 202-taxel layout across the fingertips, phalanges, and palm, so human and robot observations correspond by construction. 
Table~\ref{tab:glove-comparison} surveys previously reported interfaces, their sensing coverage, and their support for human and robot embodiments.

\textbf{Human-robot co-training for policy learning.} 
Prior work incorporates human demonstrations either as action targets, using tracked or retargeted human motion~\cite{kareer2025egomimic,kareer2025emergence,qiu2025humanoid}, or through auxiliary prediction objectives during co-training, such as predicting future frames~\cite{wu2024unleashing}, future end-effector trajectories~\cite{lepert2025masquerade}, or 3D scene flow~\cite{li2026egowam}.
Language prediction and cross-modal objectives also provide supervision for policy learning and the integration of additional sensing modalities~\cite{pi05,jones2025beyond}.
In visuo-tactile policy learning, prior work incorporates human demonstrations by aligning human and robot tactile representations using paired demonstrations~\cite{unitachand} or automatically constructed pseudo-pairs~\cite{tactalign2026}.
Unlike prior work, we combine full-hand tactile sensing with language-supervised
representation learning.
Our shared taxel layout allows human and robot observations to use a common tactile encoder, while task-phase labels provide supervision without requiring observation pairs across the two datasets.
Human demonstrations contribute to this auxiliary objective without motion retargeting or action labels.
Our experiments demonstrate improvements over robot-only training.
\section{The \name~Sensor}

We describe our \name~sensor design (Section~\ref{sec:sensor}), our visuo-tactile dexterous manipulation data collection setup, which supports human and robot demonstration collection (Section~\ref{sec:collection}), and our policy learning architecture, with a shared tactile encoder across embodiments (Section~\ref{sec:arch}).

\subsection{Sensor Design and Fabrication}\label{sec:sensor}
\begin{figure}[t]
    \centering
    \vspace{4pt}
    \begin{minipage}{0.95\linewidth}
        \setlength{\parindent}{0pt}
        \setlength{\parskip}{0pt}
        \newcommand{\hardwarepanel}[3][]{%
            \begin{minipage}[t]{0.055\linewidth}
                \vspace{0pt}\textbf{(#2)}
            \end{minipage}%
            \begin{minipage}[t]{0.945\linewidth}
                \vspace{0pt}
                \includegraphics[width=\linewidth,#1]{#3}
            \end{minipage}%
            \par\nointerlineskip
        }
        \hardwarepanel{a}{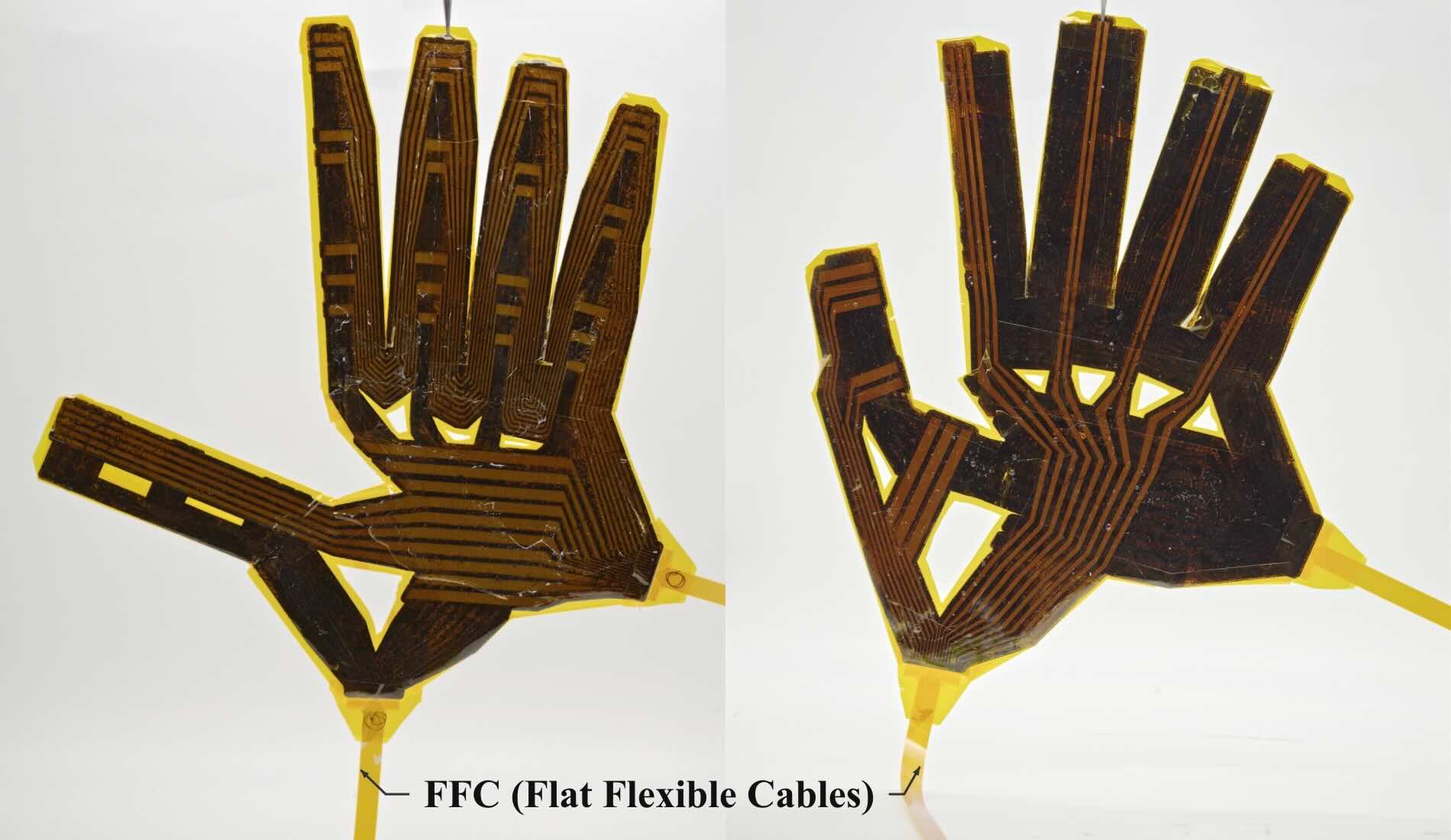}
        \vspace{0.6mm}
        \hardwarepanel{b}{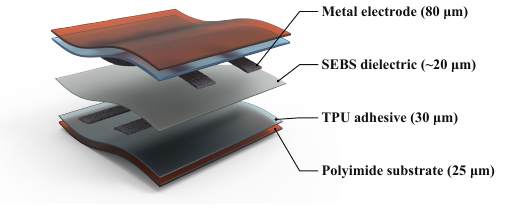}
        \vspace{0.6mm}
        \hardwarepanel{c}{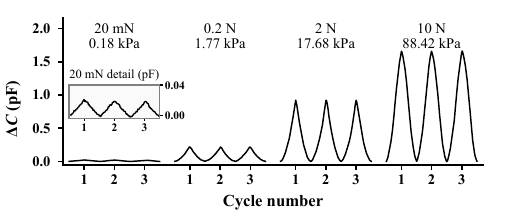}
        \vspace{0.6mm}
        \hardwarepanel{d}{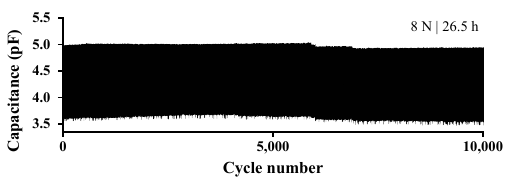}
    \end{minipage}
    \captionsetup{font=footnotesize}
    \caption{\textbf{\name~tactile skin hardware.}
    (a) Robot (left) and human (right) skins with custom-fit form
    factors and matched region-level taxel layouts.
    (b) Exploded view showing stack-up of the multilayer capacitive sensor.
    (c) Sensor responses over three consecutive cycles per load from
    20\,mN to 10\,N; boxed inset shows the
    20\,mN detail.
    (d) Sensor capacitance over 10,000 loading-unloading cycles with a peak force of 8\,N. The dense band's upper and lower edges represent loaded and unloaded sensor outputs; the approximately rectangular band indicates stable cyclic response.}
    \label{fig:sensor_hardware}
    \vspace{-10pt}
\end{figure}

We build tactile skins for both the human demonstrator and the robot hand. The devices are mechanically customized to each embodiment but share the same capacitive sensing principle, taxel size, and taxel count. Taxel spacing varies with hand geometry while preserving region-level semantic correspondence across embodiments.

In our current design, the skin provides a dense $202$-taxel pressure map with $40$ taxels assigned to the thumb, $24$ taxels to the index finger, $16$ taxels to each of the middle, ring, and pinky fingers, with the remaining taxels distributed across the palm. Each device, shown in Fig.~\ref{fig:sensor_hardware}(a), is custom-fit to its target hand by adjusting the overall geometry and taxel spacing while preserving taxel size and taxel-to-segment alignment.

The thin-film skins are fabricated using a laser-enabled metal-fabric-electrode patterning process and adopt the multilayer capacitive stack shown in Fig.~\ref{fig:sensor_hardware}(b). An $80~\mu$m-thick conductive fabric (MOS Equipment) is heat-pressed onto the $25~\mu$m-thick polyimide support layer (McMaster-Carr) using a $30~\mu$m-thick iron-on TPU film (BENECREAT). The conductive surface is then selectively ablated by UV laser to define electrically isolated traces for both the top and bottom electrode layers. A sandpaper-molded SEBS dielectric layer is sandwiched between the two patterned electrode layers to form a parallel-plate piezocapacitive sensing stack. After alignment, the multilayer device is sealed along the edges using Kapton tape and a soft SEBS layer is cast over its outer surface. Electrode traces are routed to the readout board through customized flexible flat cables (JLCPCB). The complete fabrication workflow takes approximately six hours of laser time, followed by two hours of manual assembly.

The $202$ taxels are sequentially scanned by a compact readout board measuring approximately $39.6~\mathrm{mm} \times 36~\mathrm{mm}$. The same board is used across embodiments, with wired serial communication for the robot and wireless Wi-Fi streaming for the human-worn glove. An ESP32-S3 (Espressif Systems) coordinates taxel multiplexing and data acquisition. Taxels are selected through 74HC4067 analog multiplexers (Nexperia) and measured using a PCAP04 capacitance-to-digital converter (ScioSense). With an \(850\,\mathrm{mAh}\) LiPo battery mounted beneath
the board, the glove supports approximately seven hours
of untethered operation at \(\sim30\,\mathrm{fps}\).

\name~exhibits clear capacitive responses to applied loads ranging from \(20\,\mathrm{mN}\) to
\(10\,\mathrm{N}\) in Fig.~\ref{fig:sensor_hardware}(c).
At a peak force of \(10\,\mathrm{N}\), the mean per-cycle
maximum hysteresis was \(11.05 \pm 0.13\%\) across three independently fabricated skins three months apart, lower than the \(17.5 \pm 2.8\%\) reported for STAG~\cite{sundaram2019learning}.
Separately, we evaluate cyclic stability under repeated loading at a peak force of \(8\,\mathrm{N}\) in Fig.~\ref{fig:sensor_hardware}(d).
Following 10,000 loading-unloading cycles spanning
\(26.5\,\mathrm{h}\), the sensor retained
\(97.31\%\) of its initial response span,
with endpoint peak and baseline capacitance drifts sized
\(-3.04\%\) and \(-0.35\%\) of the full-scale output
(FSO). The maximum absolute peak and baseline capacitance drifts
were \(4.86\%\) and \(9.85\%\) of the FSO, respectively.
Durability under repeated joint flexing and robustness to motion
artifacts are evaluated in Sec.~\ref{sec:durability_motion}. An appendix summarizing lessons learned during prototyping,
specifically design choices that were not robust
to repeated flexing, will be available on the website.

The current design costs less than \$10 in skin materials
and approximately \$30 for the readout circuitry at an order quantity of 100 units. We are committed to open-sourcing the sensing stack to support scalable tactile data collection across embodiments and institutions.

\subsection{Data Collection Setup}\label{sec:collection}

\begin{figure}[ht]
\vspace{4pt}
    \centering
    \includegraphics[width=\linewidth]{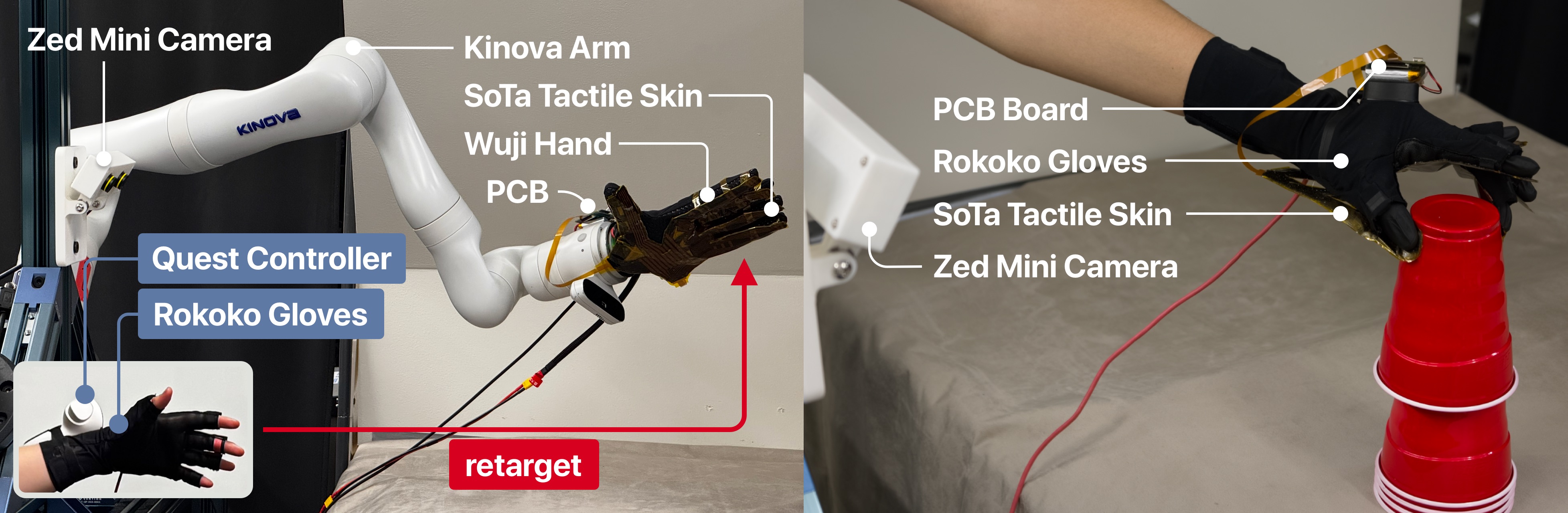}
    \captionsetup{font=footnotesize}
    \caption{\textbf{Our data collection setup.}
    (a)~\textbf{Robot teleoperation.} A Meta Quest controller tracks the operator's wrist pose and a Rokoko smart glove tracks per-finger joints; both streams are fused at 30\,Hz and command the dexterous Kinova~+~Wuji hand setup. A front-facing ZED Mini camera and our tactile skins record synchronized observations.
    (b)~\textbf{Human demonstration collection.} The operator wears a Rokoko glove and our human-fit tactile skin while performing the same tasks without the robot. Human hand configurations were recorded for an exploratory hand pose prediction objective but are not used in our method or the reported experiments.}
    \label{fig:collection_setup}
    \vspace{-10pt}
\end{figure}

The \name~data collection pipeline records synchronized visual and tactile observations for both human and robot demonstrations, together with robot proprioception and actions for robot demonstrations. The human collection setup is portable, supporting demonstration collection outside the laboratory. An illustration of our data collection system is visualized in Figure~\ref{fig:collection_setup}.

\textbf{Robot demonstrations.}
To collect teleoperated robot demonstrations, a Meta Quest headset placed on the table tracks the operator's wrist pose, while a Rokoko smart glove tracks per-finger joint positions across all five digits of each hand. Operators start and stop episodes through a lightweight command-line interface. The teleoperation interface records images, tactile signals, and wrist poses at 30\,Hz. Finger positions feed an optimization-based hand retargeter~\cite{qin2023anyteleop} that resolves them into robot hand joint angles, while wrist pose targets are mapped to arm joint angles via inverse kinematics~\cite{carpentier2019pinocchio}, with low-pass filtering applied to both for smoothness.
At each step, we record the left image stream of a front-mounted ZED Mini camera at 30\,Hz (no stereo and no wrist-mounted view), full robot proprioceptive state, and readings from our tactile skins (202 taxels) at 30\,Hz.

\textbf{Human demonstrations.}
To record human demonstrations, the operator wears a fabric glove with a human-fit tactile skin attached to the glove. The same ZED Mini camera used in robot demonstrations captures the demonstrations. We explored using Rokoko glove joint angles for an auxiliary hand pose prediction loss, but empirically found that it does not help.

We log the same observation streams as in the robot setup at 30\,Hz: a video stream from the ZED Mini camera and readings from our human-fit tactile skins.
Because our tactile layout is shared across embodiments, human demonstrations enter the training stream alongside robot demonstrations with no per-sensor alignment step.

\textbf{Synchronization and post-processing.}
All vision, proprioceptive, and tactile streams are timestamped at capture using a shared wall clock and stored per episode together with task metadata.
For each demonstration, we apply the following per-taxel
post-processing pipeline: (1) subtract a baseline computed as the mean of the first 30 frames, and set those baseline frames to zero; (2) suppress spikes in the baseline-subtracted signal; (3) normalize by the baseline and apply an exponential transformation; and (4) linearly rescale the normalized
values from $[0.1, 2.0]$ to $[0, 1]$, clipping values outside this range.
Implementation details are documented alongside the released code.

\subsection{Policy Learning Architecture}
\label{sec:arch}

We build our visuo-tactile policy on FTP-1~\cite{ftp}, a generalist
tactile manipulation policy pretrained on approximately
$3{,}000$ hours of human and robot demonstrations spanning
$21$ tactile sensors.
FTP-1 combines a PaliGemma vision-language backbone, a tactile
transformer expert, and a flow-matching action expert.
Sensor-specific tactile encoders map tactile measurements into
a shared, function-area-based token representation.
We initialize the shared model components from the pretrained
FTP-1 checkpoint and train new tactile encoders for our skins
from scratch.
Human and robot skins share these encoders and the same
functional-area representation.

\begin{figure}[t]
\vspace{4pt}
\captionsetup{font=footnotesize}
    \centering
    \includegraphics[width=\columnwidth]
        {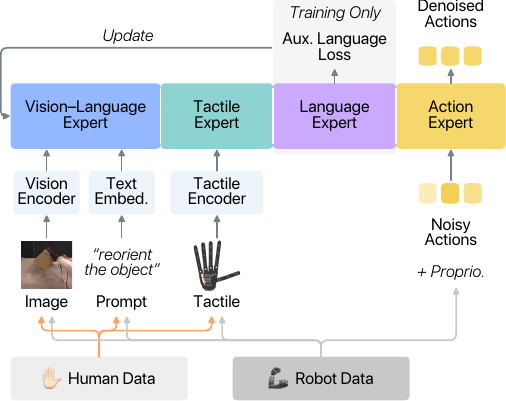}
    \caption{\textbf{Policy architecture built on FTP-1.}
    Image, prompt, and tactile observations condition the action
    expert, which generates actions through flow matching.
    A training-only language expert predicts task-phase labels
    from visual-language and tactile features.
    The action expert does not attend to the language expert,
    so the auxiliary branch is omitted at inference.}
    \label{fig:arch}
    \vspace{-8pt}
\end{figure}

We introduce a training-only language expert that predicts the
current episode-segment annotation from visual-language and
tactile features.
This branch attends to the vision-language and tactile
representations, but not to the action branch.
Its predictions are supervised with cross-entropy over the
pretrained language vocabulary, with gradients stopped through
the vocabulary projection weights.
The action expert does not attend to the language expert.

Segment annotations are obtained by manually assigning task phase
phrases to frame intervals in each demonstration via an intuitive web-based annotation interface.
For example, box reorientation uses ``approach the box,''
``reorient the box,'' and ``release.''
These phrases are mapped to canonical phase tokens and expanded
into per-frame targets.
Unlike the input task prompt, which specifies the overall goal,
the segment annotation describes the current phase of execution.
Segment labels are supervision targets only and are not provided
as policy inputs or required at inference.

The full co-training procedure has two stages.
First, we perform language-supervised representation pretraining
on human and robot demonstrations without action supervision.
This stage updates the visual encoder and its projection,
the tactile encoder and expert, and the auxiliary language branch,
while keeping the action branch and PaliGemma language
transformer frozen.
Second, we initialize policy learning from this checkpoint and
jointly optimize the FTP-1 flow-matching objective on robot
demonstrations and the auxiliary language objective on both
human and robot demonstrations.
Human demonstrations never contribute to the action loss.
The auxiliary language loss has weight $1.0$ in both stages.
\section{Experimental Evaluation}
\label{sec:result}

We evaluate three aspects of \name.
First, we test whether tactile observations improve contact-rich
manipulation relative to vision-only policies.
Second, we examine the effects of auxiliary language supervision
and human-robot co-training under a fixed robot data
budget.
Finally, we characterize sensor durability and motion-induced
artifacts.

\subsection{Experimental Setup}
\label{sec:setup}

\textbf{Tasks.}
We evaluate three contact-rich manipulation tasks:
(1) \textbf{box reorientation}, in which the robot must maintain
contact with a box and reorient it to a specified target orientation;
(2) \textbf{cup picking}, in which the robot must grasp and lift a
single thin-walled cup out of a stack without lifting more than one, dropping, or deforming; and
(3) \textbf{plug insertion}, in which the robot must align a plug
with its socket and fully insert it.
These tasks involve sustained contact, delicate grasping, and
precise alignment, respectively.

\textbf{Data Collection.}
For each task, the robot training dataset contains demonstrations
collected using a single object instance.
We use $75$ robot demonstrations each for box reorientation and
plug insertion, and $50$ for cup picking.
For each task, the human dataset contains $150$ demonstrations
spanning three training objects, one of which overlaps with the
robot-training object.

We denote robot and human demonstrations by $\mathrm{R}$ and $\mathrm{H}$, respectively.
Superscripts indicate observation modalities:
$\mathrm{V}$ denotes vision and $\mathrm{T}$ denotes touch.
The policy has access to robot proprioception in all conditions, so we omit it from the notation.
Human demonstrations contribute only to the auxiliary training objective.

\textbf{Evaluation Scenarios.}
For each task, the in-distribution (ID) condition uses the object and setup from robot demonstrations.
We additionally report one out-of-distribution (OOD) condition
for box reorientation, two for cup picking, and two for plug
insertion.
All OOD setups are absent from both human and robot training data.
We randomize the initial object positions in every evaluation episode, while ensuring that all methods within the same evaluation use the same set of randomized positions.
See Figure \ref{fig:objects} for the ID and OOD setups used in demos and evaluation.

\newcommand{\centercroppedimage}[1]{%
    \begin{tikzpicture}
        \clip (0,0) rectangle (\linewidth,\linewidth);
        \node[inner sep=0pt, anchor=center] at
            (0.5\linewidth,0.5\linewidth) {%
            \includegraphics[height=\linewidth]{#1}%
        };
    \end{tikzpicture}%
}

\begin{figure}
\vspace{4pt}
    \captionsetup{font=footnotesize}
    \centering

    \setlength{\objectsgap}{0.015\linewidth}
    \setlength{\objectsheight}{%
        \dimexpr(0.99\linewidth-2\objectsgap)*10/26\relax
    }

    \makebox[\linewidth][c]{%
        \subcaptionbox{ID objects, seen in robot and human demos\label{fig:objects_id}}[0.8\objectsheight]{%
            \resizebox{0.8\objectsheight}{\objectsheight}{%
                \clipbox{.1\width{} 0pt .1\width{} 0pt}{%
                    \centercroppedimage{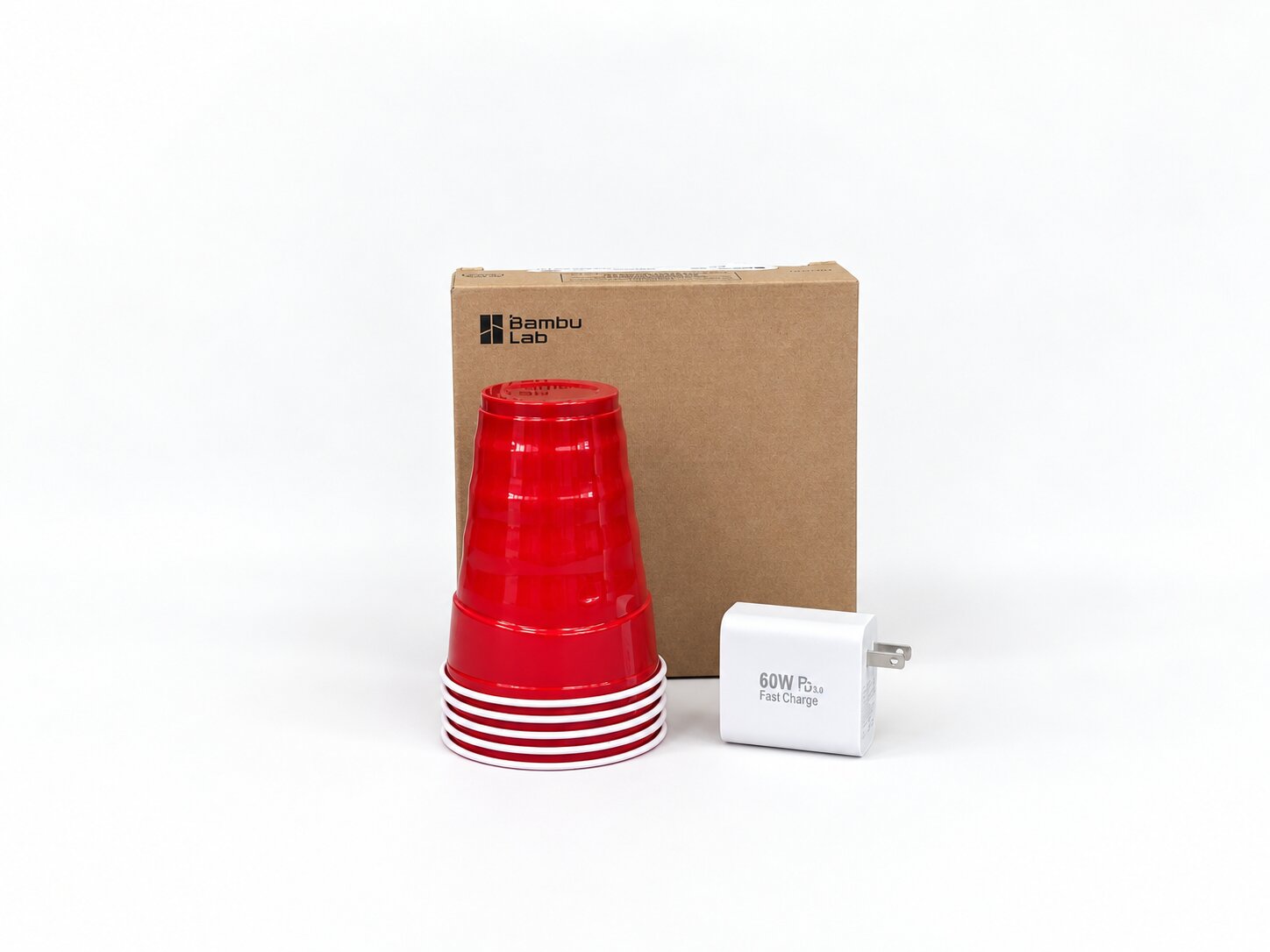}%
                }%
            }%
        }%
        \hspace{\objectsgap}%
        \subcaptionbox{Objects seen only in human demos\label{fig:objects_ood_human}}[0.8\objectsheight]{%
            \resizebox{0.8\objectsheight}{\objectsheight}{%
                \clipbox{.1\width{} 0pt .1\width{} 0pt}{%
                    \centercroppedimage{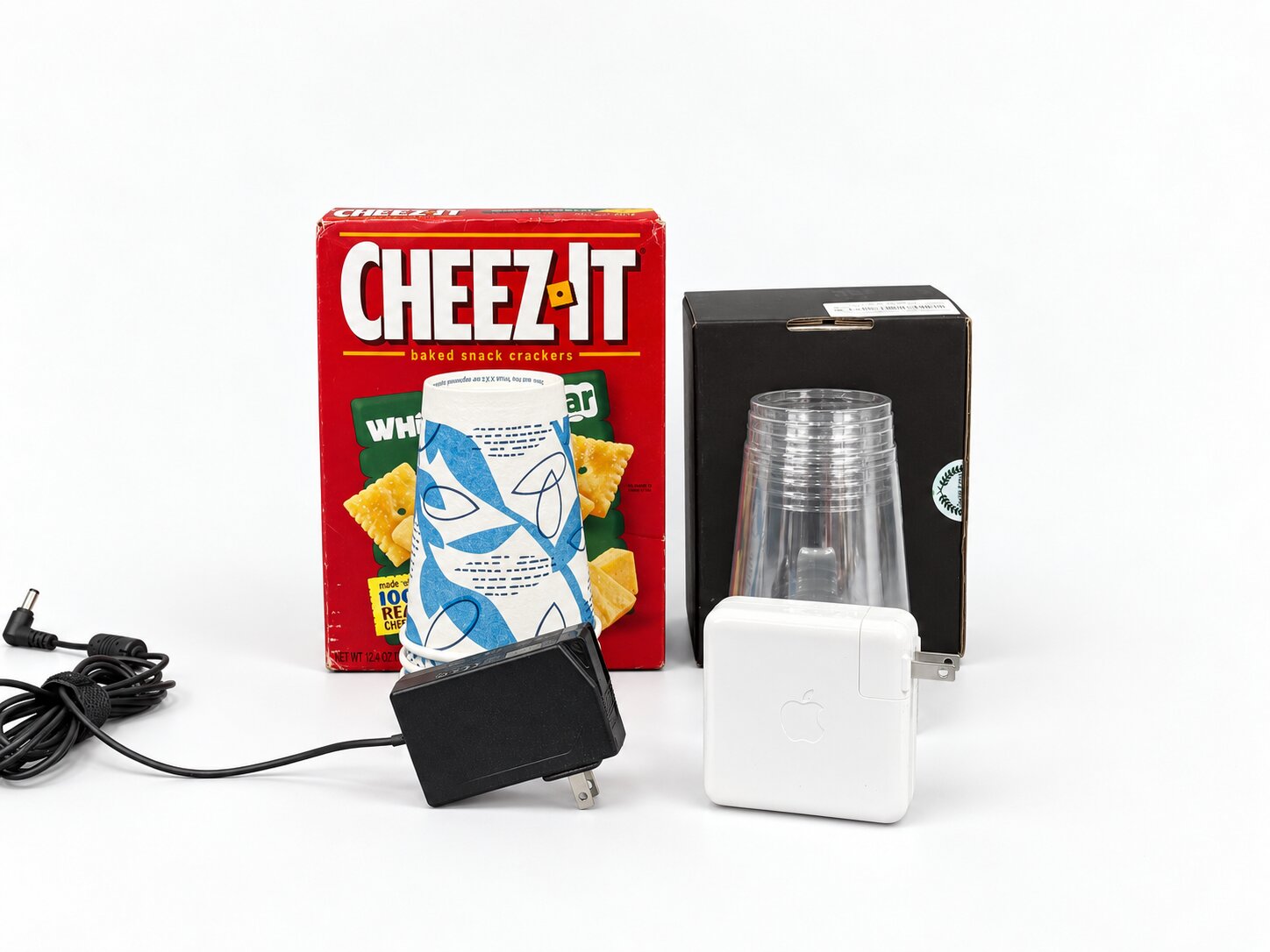}%
                }%
            }%
        }%
        \hspace{\objectsgap}%
        \subcaptionbox{Objects not seen in demos and used for OOD evaluation\label{fig:objects_ood_test}}[\objectsheight]{%
            \resizebox{\objectsheight}{\objectsheight}{%
                \centercroppedimage{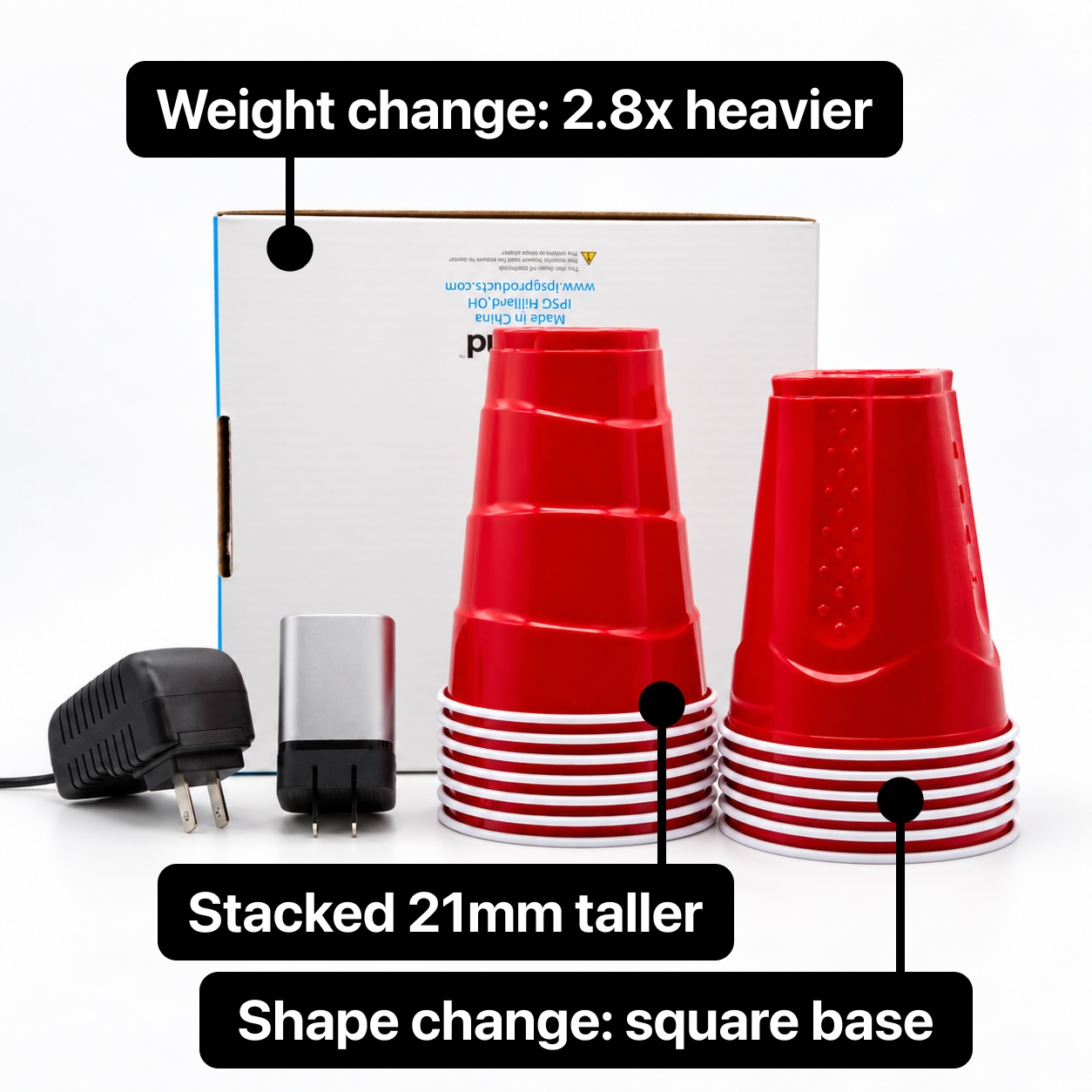}%
            }%
        }%
    }

    \caption{\textbf{Objects used in our experiments.}
    (a) In-distribution objects. We collected both robot and human demonstrations using these objects.
    (b) Objects seen in human demonstrations, but not robot demonstrations.
    (c) Objects used in OOD evaluation, which includes held-out objects and configurations.
    }
    \label{fig:objects}
    \vspace{-7pt}
\end{figure}

\textbf{Metrics.}
We report fully successful rollouts across $40$ trials for each policy, task, and evaluation condition.
Partial completions count as failures.
Given time and compute constraints, we use one training seed per method and task.

\begin{table}[t]
    \captionsetup{font=footnotesize}
    \centering
    \caption{\textbf{Training configurations in different methods.}
    All methods use robot demonstrations for action supervision.
    Two-stage training consists of representation pretraining,
    followed by joint action and auxiliary language learning.
    Human data contribute only to the auxiliary objective.
    Green ticks (\ibmtick) indicate inclusion, red crosses
    (\ibmcross) indicate absence.}
    \label{tab:baseline_comparison}
    \small
    \setlength{\tabcolsep}{4pt}
    \renewcommand{\arraystretch}{1.15}
    \resizebox{0.9\columnwidth}{!}{%
    \begin{tabular}{lcccc}
        \toprule
        \textbf{Method}
        & \shortstack{Policy\\reads tactile}
        & \shortstack{Two-stage\\training}
        & \shortstack{Human vision\\data}
        & \shortstack{Human tactile\\data} \\
        \midrule
        \rv
        & \ibmcross & \ibmcross & \ibmcross & \ibmcross \\
        \rvt
        & \ibmtick & \ibmcross & \ibmcross & \ibmcross \\
        \midrule
        \rvtlvt
        & \ibmtick & \ibmtick & \ibmcross & \ibmcross \\
        \rvthv
        & \ibmtick & \ibmtick & \ibmtick & \ibmcross \\
        \rvthvt
        & \ibmtick & \ibmtick & \ibmtick & \ibmtick \\
        \bottomrule
    \end{tabular}%
    }
    \vspace{-10pt}
\end{table}

\subsection{Effect of Tactile Feedback}

We first compare two policies initialized from the same FTP-1
checkpoint and trained on the same task-specific robot dataset:
\rv, which uses vision and proprioception,
and \rvt, which additionally receives
\name~tactile observations.
Both policies are fine-tuned directly using robot action
supervision, without representation pretraining or auxiliary
language supervision.
All other training and evaluation settings are held fixed
within this comparison.

\begin{figure}[t]
\vspace{5pt}
\captionsetup{font=footnotesize}
    \centering
    \includegraphics[width=\linewidth]{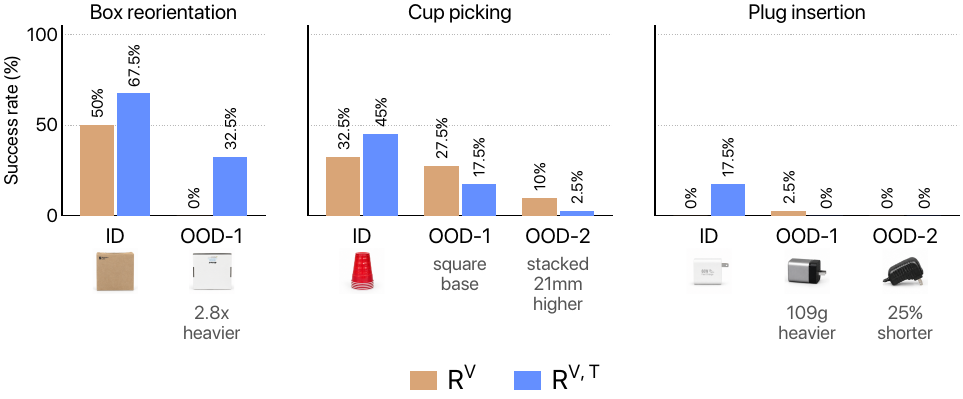}
    \caption{
        \textbf{Effect of tactile feedback.}
        We present the success rates achieved by policies trained on robot data only with and without tactile sensing. We test our policies on three tasks with in-distribution and out-of-distribution objects. The y-axis shows the percentage of fully successful rollouts out of $40$ trials. Partial completions count as failures. Results show that tactile feedback improves dexterous policy learning for in-distribution settings, while its effects in out-of-distribution settings are mixed.}
    \label{fig:results_tactile}
    \vspace{-8pt}
\end{figure}

\textit{Results.} The addition of tactile feedback improves in-distribution success on all three tasks (Figure~\ref{fig:results_tactile}) and improves box OOD success substantially. However, it does not improve OOD generalization for cup picking or plug insertion, and reduces cup OOD success relative to the vision-only policy.
We hypothesize that these differences reflect how contact patterns vary across training and evaluation objects.
Box reorientation involves broad, sustained contact that may transfer across boxes of different sizes.
Cup picking and plug insertion depend on more localized contacts that may be sensitive to object geometry.
With only one robot training object per task, policies may therefore learn tactile features that do not generalize consistently to unfamiliar objects.

\begin{figure}
\captionsetup{font=footnotesize}
    \centering
    \includegraphics[width=\linewidth]{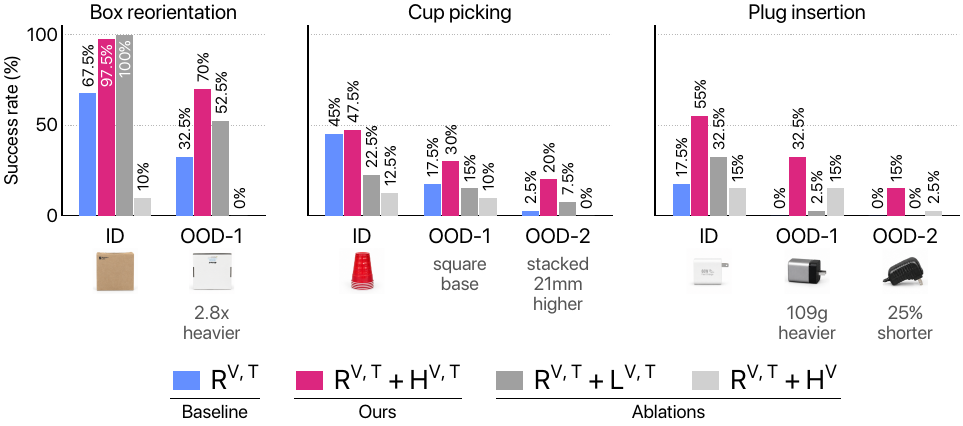}
    \caption{
        \textbf{Auxiliary supervision and human-robot co-training.}
        We present quantitative results supporting that learning dexterous visuo-tactile manipulation policies via human-robot co-training (\rvthvt) outperforms policies trained without human data (\rvt) in every reported condition, with gains concentrated in OOD settings.
        We also show ablation results on variants of our full method. \rvtlvt~denotes two-stage robot-only training with visuo-tactile language supervision. \rvthv~denotes an ablation that does not utilize tactile inputs in the auxiliary objective.
    }
    \label{fig:results_cotraining}
    \vspace{-10pt}
\end{figure}

\subsection{Auxiliary Supervision and Human-Robot Co-Training}

We also examine the contributions of auxiliary supervision and human-robot co-training.
All methods use the same FTP-1 initialization and task-specific robot demonstration dataset.
Human-robot co-training additionally uses $150$ human demonstrations per task.

Robot-only two-stage training, \rvtlvt, retains representation pretraining and visuo-tactile language supervision but excludes human demonstrations; this tests whether auxiliary training on robot data alone can reproduce the full method's gains.
Vision-only auxiliary training, \rvthv, retains human demonstrations but excludes tactile inputs from the auxiliary objective for both human and robot data; this tests whether co-training benefits from incorporating touch into auxiliary representation learning.
In this variant, the tactile encoder and expert remain frozen during representation pretraining and are updated only through robot action supervision during policy learning.
We also include the robot-only reference, \rvt, which uses action supervision without representation pretraining or auxiliary supervision.

\textit{Results.} Full human-robot visuo-tactile co-training, \rvthvt, exceeds the robot-only reference, \rvt, in every reported condition (Figure~\ref{fig:results_cotraining}), increasing mean success from 22.8\% to 45.9\%, with the largest relative gains OOD.
The two ablations examine the roles of human demonstrations and tactile auxiliary supervision.

Robot-only two-stage training, \rvtlvt, does not reproduce the full method's OOD gains: improvements over the reference are inconsistent across tasks, and cup ID performance decreases.
The full method achieves higher success than this variant in all five OOD conditions, supporting the benefit of adding human demonstrations to visuo-tactile auxiliary training.
Vision-only human co-training, \rvthv, underperforms both the full method and the reference in most conditions. Thus, additional human data with vision-only auxiliary supervision does not reproduce the full method's gains.
We use the same input modalities for auxiliary phase prediction across human and robot data, keeping the auxiliary training setup consistent across embodiments. Accordingly, this ablation evaluates tactile auxiliary supervision as a whole; it does not isolate the contribution of human touch, which would require an asymmetric modality configuration.

\textit{Analysis.} Co-training improves OOD performance where robot-only training struggles, consistent with human demonstrations broadening the training distribution's object variation relative to the single-instance robot dataset.
The human dataset spans three objects per task, one overlapping with the robot-training object, compared to the robot dataset's single instance.
This broader coverage at training time may have helped policies
generalize beyond the training objects and configurations.

\subsection{Tactile Sensor Durability and Motion Robustness}
\label{sec:durability_motion}
We compared \name's fabric-electrode stackup with three representative conductor options for wearable sensors: \(0.254\,\mathrm{mm}\) Dupont jumper wires, stencil-printed silver ink (ACI Materials SS1109) on TPU (Dingzing), and copper-polyimide traces (Pyralux AP8525R, Qnity), representative of FPCB-based gloves~\cite{murphy2025fits, opentouch2025}.
We characterized four traces per thin-film condition and four jumper wires under repeated folding. Thin-film traces were
\(0.4\,\mathrm{mm}\) wide with \(100\,\mathrm{mm}\) between contact pads. To impose more demanding test conditions, we deliberately omitted the added encapsulation used in the sensor assembly from all thin-film samples, including \name, although it was used in the sensor assembly; jumper wires retained their insulation. Samples were secured across two joints of the Wuji index finger using Kapton tape and repeatedly cycled between extension and tight fist closure. This motion folded the traces back onto themselves, imposing more severe bending than expected during typical use. Resistance was measured with the finger extended.

\begin{figure}[t]
\vspace{4pt}
\captionsetup{font=footnotesize}
    \centering
    \includegraphics[width=\linewidth]{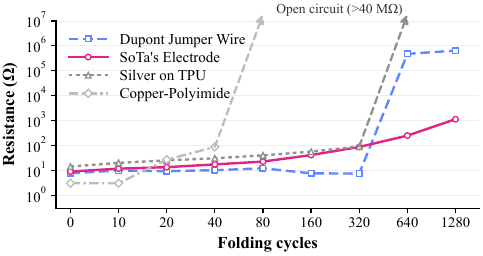}
    \caption{Conductor durability during repeated tight-fist folding on a Wuji hand. Curves show mean resistance of four conductors. Dashed arrows indicate the first open-circuit reading (\(>40\,\mathrm{M}\Omega\)) in any trace, observed at 80 cycles for copper-polyimide and 640 cycles for silver on TPU.
    \name's electrode and jumper wires reach the final checkpoint
    at 1,280 cycles with finite resistance.}
    \label{fig:trace_durability}
    \vspace{-14pt}
\end{figure}

All four \name{} traces retained continuity through 1,280 cycles (Fig.~\ref{fig:trace_durability}). Mean resistance increased from \(9.1\,\Omega\) initially to \(1.14\,\mathrm{k}\Omega\) at the final checkpoint. The final value remained below the readout's nominal \(10\,\mathrm{k}\Omega\) discharge resistance and was approximately 1/560 of the jumper wires (\(643\,\mathrm{k}\Omega\)).
A fresh baseline is acquired each time the glove is worn. Under the RC readout model, baseline subtraction followed by normalization to that baseline cancels resistance-induced scaling of the measured capacitance. The other two thin-film conditions exhibited open-circuit failures, first observed at 80 cycles for copper-polyimide and 640 cycles for silver on TPU. See individual resistance trajectories on our website.

To assess motion-induced artifacts, four experimenters
with hand widths ranging from \(6.5\) to \(8.9\,\mathrm{cm}\)
wore the same medium-sized glove and flexed each finger
toward the palm without any external contact. We computed regional RMS of \(\Delta C/C_0\) relative to each taxel's initial no-load baseline. Across all five flexion conditions, motion-induced responses were strongest on the flexed finger, with smaller responses on the unflexed fingers and palm
(Table~\ref{tab:motion_artifacts}).
These responses were modest in magnitude: the largest mean regional RMS (\(6.18\%\)) was comparable to the normalized response of the bench-tested taxel under a light load of \(0.2\,\mathrm{N}\).

\begin{table}[t]
\vspace{5pt}
    \centering
    \captionsetup{font=footnotesize}
    \caption{Motion-induced responses during individual finger flexion.
    Values are regional RMS of $\Delta C/C_0$ (\%) over ten flex–relax cycles, reported as
    mean $\pm$ SD across four experimenters. Each \(60\,\mathrm{s}\) trial comprised a \(10\,\mathrm{s}\) still-hand baseline followed by ten cycles of \(2\,\mathrm{s}\) flexion and \(3\,\mathrm{s}\) relaxation; RMS was computed over the \(50\,\mathrm{s}\) after the baseline. The unflexed-finger region excludes the flexed finger and palm.}
    \label{tab:motion_artifacts}
    \small
    \resizebox{0.9\columnwidth}{!}{%
    \setlength{\tabcolsep}{3pt}
    \begin{tabular}{lccc}
        \toprule
        Finger & \multicolumn{3}{c}{RMS of $\Delta C/C_0$ (\%)} \\
        \cmidrule(l){2-4}
        flexed & Flexed finger & Unflexed fingers & Palm \\
        \midrule
        Thumb & $3.66 \pm 0.63$ & $1.69 \pm 0.89$ & $2.30 \pm 0.87$ \\
        Index & $6.18 \pm 1.10$ & $1.54 \pm 0.35$ & $1.60 \pm 0.46$ \\
        Middle & $5.49 \pm 0.47$ & $2.55 \pm 0.48$ & $1.20 \pm 0.20$ \\
        Ring & $5.60 \pm 0.56$ & $1.97 \pm 0.62$ & $1.34 \pm 0.33$ \\
        Little & $4.09 \pm 0.86$ & $1.45 \pm 0.41$ & $1.60 \pm 0.44$ \\
        \bottomrule
    \end{tabular}
    }
    \vspace{-8pt}
\end{table}
\section{Discussion and Conclusion}
\label{sec:conclusion}

We presented \name, a low-cost, durable capacitive tactile skin that provides full-hand coverage on human and robot hands.
Our motivation is that visuo-tactile robot data is expensive to collect through teleoperation, while human demonstrations are cheap but only useful if both embodiments produce corresponding tactile signals.
\name~addresses this by fitting the skin to each hand while preserving taxel correspondence, so human and robot demonstrations share a common tactile encoder without a learned cross-sensor mapping.
Across box reorientation, cup picking, and plug insertion tasks, tactile observations improve in-distribution success.
With a fixed robot-demonstration budget, adding human visuo-tactile demonstrations improves every evaluated condition, raising average success from 22.8\% to 45.9\%, with the largest gains out of distribution. These results suggest that tactile human demonstrations are a practical way to supplement costly robot teleoperation.

Several questions remain open.
We used a fixed set of 150 human demonstrations per task, so the effects of increasing the size and diversity of the human dataset, and the extent to which human demonstrations can replace robot data, remain to be measured. We also did not compare our shared taxel layout against learned cross-sensor mapping approaches. 
Finally, our hardware tests demonstrate stability under repeated loading and small motion-induced responses, but long-term wear under sustained use remains to be evaluated.

By releasing the fabrication workflow, sensor designs, Gerber files for ordering the readout boards, and firmware, we aim to enable other research groups to reproduce and adapt the sensing platform.
Because the glove is wearable and untethered, it can collect tactile demonstrations across diverse objects, tasks, and demonstrators without a robot. Future work will study how far this scaling can go and extend the skin to other end effectors while preserving the shared taxel layout.
\section{Acknowledgements}
\label{sec:acknowledgements}

We thank all IPRL members for their valuable suggestions on manuscript writing and presentations. We also thank Mandi Zhao, Sunshine Jiang, Chengyi Xu and Junyi Zhao for valuable discussions and feedback. We would like to thank Wuji Tech for providing the robot hand hardware. We also thank Stanford Marlowe~\cite{marlowe2025} and Supermicro for providing computational resources.
HL was supported by the Stanford Undergraduate Visiting Research Program (UGVR).
AL was supported by a Wallenberg–Bienenstock Postdoctoral Fellowship.
RA was supported in part by the UK Advanced Research and Invention Agency (ARIA) under the Robot Dexterity Programme. BS, WC, and ZB acknowledge support from Yi Man Liu and Sai Fu Fung, the Tianqiao and Chrissy Chen Ideation and Prototyping Lab, seed funding from the Stanford Wearable Electronics Initiative (eWEAR), and the Stanford Robotics Center.
Part of this work was performed at nano@stanford (RRID:SCR\_026695). This work was also supported by the National Science Foundation under Grant Number 2342246.

ChatGPT, Codex, and Claude have been used to assist in code  , figure creation, and text editing.

\begingroup
\balance
\bibliographystyle{IEEEtran}
\bibliography{references}
\endgroup

\end{document}